\DocumentMetadata{}
\documentclass[sigconf]{acmart}

\usepackage{algorithm}
\usepackage{algorithmic}
\usepackage{multirow}

\AtBeginDocument{%
  \providecommand\BibTeX{{%
    \normalfont B\kern-0.5em{\scshape i\kern-0.25em b}\kern-0.8em\TeX}}}

\copyrightyear{2025}
\acmYear{2025}
\setcopyright{rightsretained}
\acmConference[SIGIR '25] {Proceedings of the 48th International ACM SIGIR Conference on Research and Development in Information Retrieval}{ July 13--18, 2025}{Padua, Italy.}
\acmBooktitle{Proceedings of the 48th International ACM SIGIR Conference on Research and Development in Information Retrieval (SIGIR '25), July 13--18, 2025, Padua, Italy}
\acmDOI{10.1145/3726302.3730018}
\acmISBN{979-8-4007-1592-1/25/07}

\makeatletter
\gdef\@copyrightpermission{
  \begin{minipage}{0.3\columnwidth}
   \href{https://creativecommons.org/licenses/by/4.0/}{\includegraphics[width=0.90\textwidth]{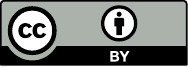}}
  \end{minipage}\hfill
  \begin{minipage}{0.7\columnwidth}
   \href{https://creativecommons.org/licenses/by/4.0/}{This work is licensed under a Creative Commons Attribution International 4.0 License.}
  \end{minipage}
  \vspace{5pt}
}
\makeatother

\begin{document}

\title{Knowing You Don't Know: Learning When to Continue Search in Multi-round RAG through Self-Practicing}


\author{Diji Yang}
\authornotemark[1]
\email{dyang39@ucsc.edu}
\affiliation{
  \institution{University of California Santa Cruz}
  \city{Santa Cruz}
  \country{USA}
}

\author{Linda Zeng}
\authornote{Equal contribution.}
\email{lindazeng979@gmail.com}
\affiliation{
  \institution{The Harker School}
  \city{San Jose}
  \country{USA}
}

\author{Jinmeng Rao}
\email{jinmengrao@google.com}
\authornote{Now at Google DeepMind.}
\affiliation{%
  \institution{Mineral.ai}
  \city{Mountain View}
  \country{USA}
}

\author{Yi Zhang}
\email{yiz@ucsc.edu}
\affiliation{%
  \institution{University of California Santa Cruz}
  \city{Santa Cruz}
  \country{USA}
}

\renewcommand{\shortauthors}{Diji Yang et al.}

\begin{abstract}
Retrieval Augmented Generation (RAG) has shown strong capability in enhancing language models' knowledge and reducing AI generative hallucinations, driving its widespread use. However, complex tasks requiring multi-round retrieval remain challenging, and early attempts tend to be overly optimistic without a good sense of self-skepticism. Current multi-round RAG systems may continue searching even when enough information has already been retrieved, or they may provide incorrect answers without having sufficient information or knowledge. Existing solutions either require large amounts of expensive human-labeled process supervision data or lead to subpar performance. 

This paper aims to address these limitations by introducing a new framework, \textbf{SIM-RAG}, to explicitly enhance RAG systems' self-awareness and multi-round retrieval capabilities. To train SIM-RAG, we first let a RAG system self-practice multi-round retrieval, augmenting existing question-answer pairs with intermediate inner monologue reasoning steps to generate synthetic training data. For each pair, the system may explore multiple retrieval paths, which are labeled as successful if they reach the correct answer and unsuccessful otherwise. Using this data, we train a lightweight information sufficiency Critic. At inference time, the Critic evaluates whether the RAG system has retrieved sufficient information at each round, guiding retrieval decisions and improving system-level self-awareness through in-context reinforcement learning.

Experiments across multiple prominent RAG benchmarks show that SIM-RAG is an effective multi-round RAG solution. Furthermore, this framework is system-efficient, adding a lightweight component to RAG without requiring modifications to existing LLMs or search engines, and data-efficient, eliminating the need for costly human-annotated mid-step retrieval process supervision data.~\footnote{All code and data are available at \url{https://github.com/ucscirkm/SIM-RAG}.}
\end{abstract}




\maketitle

\section{Introduction}
\label{sec:intro}
    \begin{figure*}
        \centering
        \includegraphics[width=\linewidth]{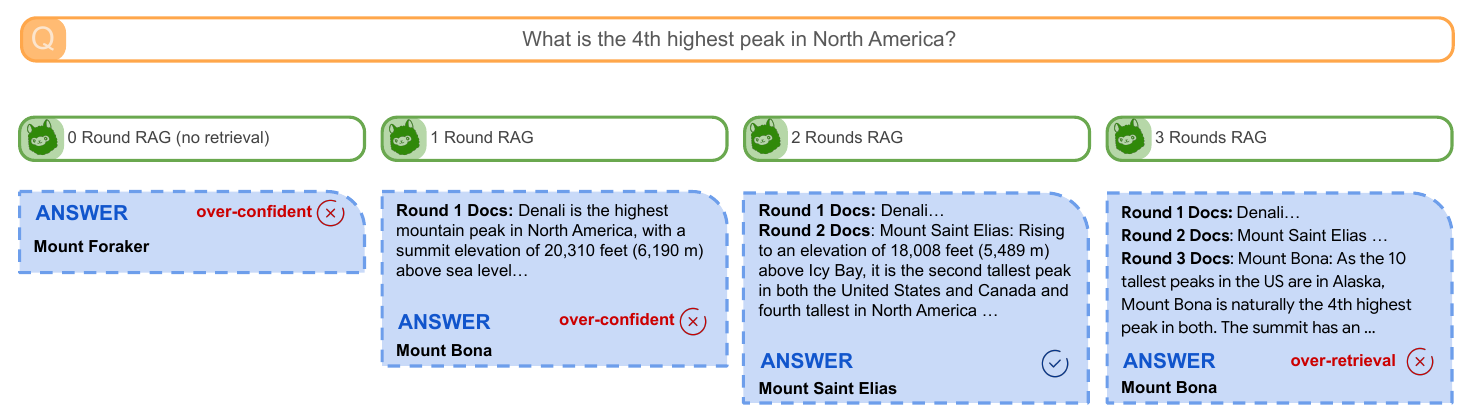}
        \caption[]{      
        A key challenge in multi-round RAG systems\setcounter{footnote}{1}\footnotemark~is determining the optimal stopping point for retrieval and then generating the answer. This figure illustrates two typical patterns that hurt performance: Over-Confidence (e.g., stopping too early, as seen in 0 and 1 Round RAG, where the system provides an incorrect answer based on limited context) and Over-Retrieval (e.g., retrieving unnecessary information in 3 Rounds RAG which introduces a long and overly complex context that confuses the LLM and leads to incorrect answers.).
        }
        \Description[]{}
        \label{fig:teaser}
    \end{figure*}

Large language models (LLMs) have shown decent results in multi-step reasoning benchmarks, such as mathematical competition~\cite{openai2024o1}, yet Retrieval Augmented Generation (RAG) systems still lag behind human performance in complex tasks that involve multi-round retrieval~\cite{tang2024multihop}. One of the major challenges in RAG is the need for strong self-awareness of its knowledge boundaries. In a closed-book reasoning setting, all knowledge is embedded inside the LLM and remains inherently static, regardless of how the problem is decomposed or how a Chain-of-Thought (CoT) is structured. In contrast, RAG involves external augmented information accessed through retrieval, potentially shifting the system’s internal knowledge boundary. Furthermore, retrieval adds additional complexity and uncertainty, which may accumulate over extended reasoning sequences in multi-round RAG systems. 

Human intelligence addresses this issue through meta-cognition (i.e., knowing when you don't know)~\cite{flavell1979metacognition,liu2024right}. 
Humans can continuously assess their knowledge boundaries and adapt their search behaviors as needed in dynamic information environments, such as when using a search engine. After reviewing retrieved results at each time point, humans assess whether sufficient information has been gathered, decide whether further search is necessary, and issue new queries to better address the current information needs.

Meta-cognition is challenging for LLMs due to their noise sensitivity and limited self-awareness of knowledge boundaries~\cite{liu2024right,huang2024large}. As illustrated in Figure~\ref{fig:teaser}, systems that rely on an LLM to decide the number of retrieval rounds make two types of errors: Over-Confidence, resulting in incorrect answers from insufficient information, and Over-Retrieval, where excessive and distracting information confuses the LLM. Thus, a core problem for multi-round RAG is \textit{knowing you don't know} so that a system can either continue searching only if it is necessary or refrain from answering when the available information is insufficient to support a credible response. 
As an underexplored problem, recent research either requires a large amount of costly human-labeled supervision data~\cite{yang2024rag} or produces suboptimal performance~\cite{asai2023self}.

\footnotetext{In this figure, the multi-round RAG refers to a system that relies on LLM to determine how many rounds of retrieval are needed, including 0, 1, or multiple rounds.}

The optimization of a RAG system often employs outcome supervision to directly align the initial input with the final output~\cite{lewis2020retrieval,guu2020retrieval}. While outcome supervision with simple question-answer (QA) pairs has proven effective for single-step RAG~\cite{borgeaud2022improving,izacard2023atlas}, when an LLM can rapidly learn to map a question to its direct answer or query, it appears inadequate for learning the optimal reasoning path in a multi-round RAG setting, when the answer or next-round query is context-dependent over rounds.

Machine Learning researchers have recently found that process supervision is a promising alternative to outcome supervision for enhancing self-awareness in complex reasoning tasks at the inference time thinking stage~\cite{kumar2024training,lightmanlet,openai2024o1}. With a well-trained reward model from human-labeled CoT data, optimization can be performed by explicitly supervising intermediate reasoning steps via parameter tuning or training-free verbal reinforcement learning (RL)~\cite{yang2024tackling,shinn2024reflexion,monea2024llms}.
An early attempt from the IR community, IM-RAG~\cite{yang2024rag}, has explored process supervision for multi-round RAG  by simulating the human Inner Monologue reasoning process (i.e., multi-round self-talk within one's mind). It optimizes each mid-step query or answer via actor-critic 
RL~\cite{NIPS1999_6449f44a}. Despite its strong performance, the training relies on expensive human-annotated supporting documents to generate multi-round reasoning and retrieval training data (i.e., information-seeking chains with labels).

The lack of labeled training data is the main challenge in widely applying process supervision to RAG. Unlike other LLM tasks, such as coding or mathematical reasoning, annotating gold reasoning chains in RAG tasks is difficult because different LLMs may have varied internal knowledge, leading to distinct information needs even in the same context. Thus, human-annotated, LLM-independent information-seeking chains may not align with an LLM's behavior and knowledge, making high-quality multi-round RAG training data costly to label. 

This work addresses the labeled data shortage problem when adapting process supervision to multi-round RAG systems. We propose \textbf{SIM-RAG} (\textbf{S}elf-practicing for \textbf{I}nner \textbf{M}onologue-based \textbf{R}etrieval \textbf{A}ugmented \textbf{G}eneration), a practical multi-round framework that can be learned through two stages. First, in the \textit{Self-Practicing} stage, we generate synthetic process supervision data by distilling the system's inner monologues and corresponding process labels. This inner monologue captures the system’s internal complex reasoning trajectory across its components and can also be interpreted as a form of dynamic reasoning chain. Unlike synthetic data generation using the strongest models, which focuses on producing near-human quality data~\cite{alpaca}, Self-Practicing generates data reflecting the given AI system's capability.
Then, during the \textit{Critic Training} stage, we use the generated data to train a Critic, which is system-specific and context-aware. When a SIM-RAG system is used at inference time, the Critic repeatedly checks the knowledge boundary based on available information and provides process supervision to optimize LLM's behavior via in-context RL~\cite{shinn2024reflexion}, alleviating Over-Confidence and Over-Retrieval problems.  
In summary, our contributions are as follows:
\begin{itemize}
    \item
    To simulate human-like meta-cognition in complex reasoning, we propose the \textbf{SIM-RAG} framework, which continuously assesses the system's knowledge boundaries and adapts its search behaviors accordingly. By using information sufficiency as a guiding principle for process supervision, SIM-RAG enhances the inference-time thinking capabilities of LLMs, enabling dynamic, multi-round retrieval and reasoning for complex tasks.
    \item 
    To address the data challenges for RAG system training, we introduce the Self-Practicing Algorithm (Algorithm \ref{alg:data_generation}). This algorithm generates synthetic training data, providing a lower-cost alternative to human-annotated supporting documents or labeled information-seeking chains, and producing training data that more accurately reflects the given AI system’s current capabilities.
    \item Our experiments on three standard benchmarks show that SIM-RAG is a lightweight and effective solution, capable of robust performance on complex reasoning tasks across diverse question-answering scenarios. 
\end{itemize}

\section{Related Work}

\subsection{Retrieval Augmented Generation}
Retrieval Augmented Generation (RAG) enhances large language models (LLMs) by retrieving external knowledge during inference, addressing limitations such as reliance on fixed pre-trained knowledge bases and susceptibility to hallucinations~\cite{gao2023retrieval}. Pre-trained LLMs often lack up-to-date or domain-specific information, while hallucinations arise when the model generates plausible but incorrect content. By incorporating external retrieval, RAG enables more accurate and grounded responses.
In standard RAG systems, also known as single-round RAG, the retrieval process involves using the user’s question or an LLM-generated query to search a knowledge base~\cite{izacard2023atlas,borgeaud2022improving,ram2023context}. These systems are effective for tasks with straightforward information needs, where the required information is fully available in a single retrieval step and does not depend on iterative reasoning or multiple rounds of interaction.

However, many real-world tasks involve dynamic and complex reasoning, where the required information cannot be retrieved in a single step. For instance, answering a question may require synthesizing information from multiple documents, clarifying ambiguities, or filling gaps in the initial retrieval. In such cases, single-round RAG systems fall short, as they lack mechanisms to iteratively refine their retrieval and reasoning strategies~\cite{trivedi2023interleaving}. This has motivated the development of multi-round retrieval and reasoning systems.

    \begin{figure*}
        \centering
        \includegraphics[width=0.98\linewidth]{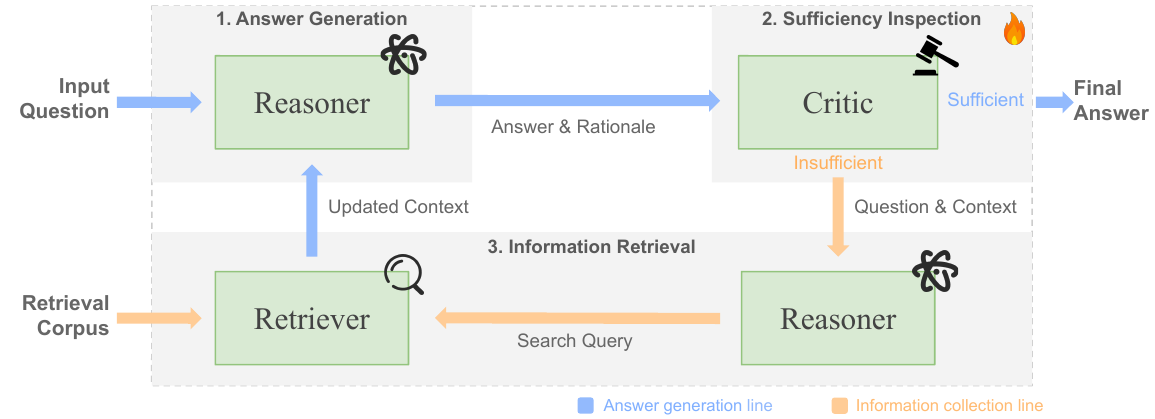}
        \caption{SIM-RAG overview at inference time, consisting of three main components: (1) Answer Generation, where the Reasoner generates an answer and rationale; (2) Sufficiency Inspection, where the Critic decides whether the generated answer to accept the answer or trigger refinement; and (3) Information Retrieval, where, if deemed insufficient, a query is generated to retrieve additional documents from the corpus to update the context. This iterative process continues until the Critic deems the answer sufficient or the maximum number of retrieval rounds is reached. The orange line represents the information collection path, while the blue line represents the answer generation path, visualizing the information flow between components.}
        \Description[High-level structure of SIM-RAG.]{This figure provides an overview of the SIM-RAG architecture, showing its key components and flow of information.}
        \label{fig:overview}
    \end{figure*}
    
\subsection{Multi-Round RAG}
\label{sec:mutliRoundRAG}
Multi-round RAG has shown great potential for tackling tasks involving dynamic and complex reasoning, where iterative interactions with external knowledge sources are necessary to refine responses. However, a core challenge in multi-round RAG is determining information sufficiency—deciding when the retrieved information is adequate to answer the query or if further retrieval steps are required~\cite{su2024dragin,yang2024rag}.
Existing works on multi-round RAG have explored various techniques to address this challenge.

\paragraph{Training-free system}
Training-free methods have gained popularity due to their flexibility and ease of deployment, as they do not require task-specific optimization and can be seamlessly integrated into existing pipelines. These systems rely on the inherent capabilities of LLMs to determine when retrieval should stop. One approach involves reflection-based self-critique~\cite{shinn2024reflexion,trivedi2023interleaving}, where the model evaluates its own knowledge through carefully designed prompts or in-context learning~\cite{wang2023self,wang2024llms}. This technique leverages the strong pre-trained knowledge of LLMs and allows the use of any LLM as a reasoning backbone. However, it is inherently constrained by the limitations of the underlying model, including a tendency to generate hallucinations or over-confident but incorrect responses~\cite{yang2024reinforcing,stechly2023gpt}.
Some methods take advantage of the internal states of LLMs, such as token-level attention weights~\cite{su2024dragin} or confidence scores~\cite{yao2024seakr}, to decide retrieval sufficiency. These signals can offer insights into the model’s reasoning process, but often require access to model weights and, consequently, cannot be used with closed-source LLMs. Additionally, the lack of interpretability in latent representations undermines trustworthiness, making these signals less suitable for applications~\cite{xiong2024can} such as healthcare, where trustworthiness plays a critical role. 

Recent studies have explored the use of well-trained models, such as GPT-4, as reward models~\cite{nguyen2024reward}. However, this approach remains constrained by the inherent pre-training biases of the reward model due to the absence of task-specific optimization~\cite{stechly2024self}. 
More broadly, the lack of training presents a double-edged sword: while it simplifies deployment and enhances ease of use, it simultaneously restricts the potential for further performance improvement

\paragraph{Learnable framework} Several learning frameworks for multi-round RAG have been proposed in the past two years. 
A recent work trains a classifier to categorize the difficulty of user queries into three classes ~\cite{jeong2024adaptive}, and applies different retrieval strategies to each class: no retrieval for simple queries, single-step retrieval for medium-difficult queries, and IR-CoT~\cite{trivedi2023interleaving} for complex queries. Since this approach primarily focuses on choosing different retrieval strategies, it is not directly comparable to our work. Instead, we used IR-CoT as one of our baselines.
Self-RAG~\cite{asai2023self} is inspired by the concept of Reinforcement Learning with Human Feedback (RLHF)~\cite{ouyang2022training}. It first trains a separate critic model using high-quality data to evaluate information sufficiency and then fine-tunes the LLM model in full size, equipping it with self-critique capabilities. Moreover, instead of standard online RLHF, Self-RAG uses outcome supervision~\cite{lightmanlet} to generate sequences of text that include special tokens to trigger retrieval operations. 
Although this method shows potential, its substantial data and computational costs limit its practicality for deployment and domain-specific adaptation~\cite{yan2024corrective}. 

Recently, the Machine Learning community has shown that process supervision~\cite{lightmanlet}, which supervises intermediate reasoning steps, can significantly improve multi-step reasoning tasks compared to outcome supervision~\cite{kumar2024training}. Motivated by these findings, IR researchers proposed IM-RAG~\cite{yang2024rag}, 
a framework that employs reinforcement learning via Proximal Policy Optimization (PPO)~\cite{schulman2017proximal} to jointly optimize the reasoning chain and retrieval queries (i.e., learning the model’s inner monologue) in multi-round RAG. IM-RAG has shown notable performance improvements.
However, the method faces challenges in broader applicability due to its reliance on costly annotated support documents to define dataset-specific rule-based reward functions, and it lacks a principled mechanism for determining when to terminate the retrieval process.

\subsection{LLM Self-Training for Complex Reasoning}
With the increasing need for fresh data, recent advancements in post-training have shifted the focus toward using model-generated data to improve reasoning (i.e., self-training)~\cite{zelikman2022star,hosseini2024v,singhbe2024yond}. In this paradigm, LLMs generate multiple outputs for a given input and use a reward signal to identify and train on high-quality samples~\cite{zhang2024rest}. This iterative process enables models to improve their reasoning capability without relying exclusively on human-labeled data.

Adapting self-training to RAG has some new complications compared with pure language tasks (e.g., commonsense reasoning), which rely on static knowledge embedded within an LLM. In a multi-round RAG setting, newly retrieved information changes the knowledge boundary at each turn, the additional information can either support or hinder subsequent reasoning, and errors in early retrieval can propagate through later stages due to the sensitivity of LLMs to noisy contexts~\cite{petroni2020context}. Optimizing self-awareness in RAG systems requires determining whether the current knowledge is adequate (when to retrieve) and issuing effective queries to acquire additional information that meets the current needs (what to retrieve). Because the knowledge boundary changes over turns and noise is added, multi-round RAG requires a higher level of self-awareness in the cognitive hierarchy~\cite{liu2024right} compared to multi-step reasoning tasks with static knowledge. This work tries to solve these unique challenges RAG systems face during self-training. 

Practitioners have used LLM for self-critique when self-training, internalizing the critique capability. However, the research community is questioning whether LLM truly has enough self-awareness to satisfy self-critique~\cite{huang2024large,stechly2023gpt,stechly2024self}. 
Motivated by these findings, we choose to use an external Critic rather than self-critique. SIM-RAG employs a lightweight Critic that remains unattached from the LLM and is trained on a single task: checking the information sufficiency.

\section{Methodology}

This section introduces the SIM-RAG framework, outlining its design in Section~\ref{sec:sim-rag-main}. We then provide an in-depth explanation of two core stages for framework training: Self-Practicing for inner monologue distillation and labeling (Section~\ref{sec:data-gen}) and Critic training (Section~\ref{sec:training}). Finally, we elaborate on the overarching rationale for the framework’s design and its inference methodology (Section~\ref{sec:inference}).

\subsection{SIM-RAG}
\label{sec:sim-rag-main}
The overview of the SIM-RAG framework is illustrated in Figure~\ref{fig:overview}, following the information flow during the inference-time thinking process. 
The system comprises three main components: (1) a Reasoner (an LLM) that generates queries or provides answers based on the context; (2) a Retriever (e.g., a search engine) that retrieves documents given the Reasoner’s query; and (3) a learnable Critic (a lightweight discriminative model) that identifies when the Reasoner’s current knowledge and reasoning are insufficient. 
As a fully modularized system, SIM-RAG organizes these components into three functional modules, described as follows in the order they are used during inference.

\paragraph{Answer Generation}

In this stage, the Reasoner (LLM) receives the initial question or task $Q$ from the user and any available context $C$ from previous retrieval steps (an empty string at turn 0). The Reasoner produces an answer $A'$ and a corresponding rationale $r$. 
Although the context is initially empty, the Reasoner benefits in subsequent turns from an accumulated context containing prior search queries and retrieved documents. 
The Reasoner’s objective at this stage is to produce its best-guess answer based on all information currently available.
Thus, the model chosen for the Reasoner can be any language model capable of answering questions, provided it can generate an answer and a rationale in natural language that can later be evaluated by the Critic.

\paragraph{Sufficiency Inspection}
In complex reasoning problems, humans can continuously assess whether they have sufficient information and a correct answer as they progress through a long reasoning chain. This ability, known as meta-cognition, enables individuals to monitor and support their reasoning throughout the thinking process.

SIM-RAG employs a Critic to simulate a similar meta-cognitive function. After receiving the Reasoner’s proposed answer-rationale pair $(A',r)$, the Critic examines them alongside the initial question $Q$ and the retrieved documents contained in the current context $C$. 
If the Critic determines that the answer $A'$ is adequately supported by the evidence from $(Q,C,r)$, the system returns $A'$ as the final answer to the user. If the Critic judges $A'$ to be insufficient, due to lack of information, inadequate support from retrieved data, inconsistencies with known facts, or similar issues, the system discards the current attempt and proceeds to the Information Retrieval module. This design helps prevent the propagation of flawed reasoning paths by ensuring that only well-supported answers are returned to the user.

\paragraph{Information Retrieval}
After the Critic determines that the Reasoner is unable to answer the question based on all available information, the system triggers the Information Retrieval module. 
The Reasoner generates a search query $q$ based on the user’s question and the current context. 
This query is then passed to the Retriever, which returns the most relevant external knowledge.
Both the search query and the returned documents are appended to the $C$, which will be fed into the next round of Answer Generation. By integrating newly retrieved information, the Reasoner may be better equipped to converge on a well-supported answer in subsequent iterations.
Notably, the Reasoner in this stage is flexible and could be the same or a different LLM used in the Answer Generation block. For simplicity, and without loss of generality, we used the same LLM to generate both queries and answers in our experiments. However, in practice, LLMs optimized for each functionality may result in better performance, as developing a good answer may require different abilities from issuing a good query~\cite{yang2024rag}. 

\paragraph{Iterative Framework}
After updating the context $C$, the system loops back to the Answer Generation stage, with the newly retrieved information helping to expand the Reasoner's knowledge boundary. This iterative process, consisting of three steps per iteration, simulates a human-like search and reasoning loop that continually reevaluates the adequacy of its current explanation and dynamically seeks additional information as needed. The Answer Generation, Sufficiency Inspection, and Information Retrieval steps repeat until the Critic determines the answer is sufficiently grounded or the maximum number of iterations is reached to prevent an infinite loop. This cyclical, meta-cognitive design aims to maximize the correctness and completeness of the final response.

\subsection{Self-Practicing}
\label{sec:data-gen}

    \begin{algorithm}
    \caption{Self-Practicing Algorithm.
    \textbf{Description.} This algorithm generates labeled inner monologue data by enabling the RAG system to perform self-practice. It automatically searches, attempts to generate answers, and checks whether the generated answer for each sequence of actions is correct.}
    \label{alg:data_generation}

    \begin{algorithmic}[1]
    \STATE \textbf{Notation:}
    
    \STATE $\mathcal{IR}$: Information Retriever
    \STATE $\mathcal{R}$: Reasoner LLM
    
    \STATE $Q$: Question
    \STATE $A$: Ground-truth Answer
    \STATE $C$: Empty list to track the context
    \STATE $q_t$: Search query from LLM at the $t$-th turn
    \STATE $d_t$: Retrieved document at the $t$-th turn
    \STATE $r$: Rationale
    \STATE $A'$: LLM generated answer
    \STATE $T$: Maximum number of turns allowed
    
    \FOR{each data point}
        \STATE $C \gets \emptyset$, ${i} \gets 0$ \label{alg:line-initialization} \hfill \textcolor{gray}{// initialize context and turn counter}
        \WHILE{{t} < {T}}
            \STATE $(A', r) \gets \mathcal{R}(Q, C)$ \label{alg:line-generateanswer}  \hfill \textcolor{gray}{// generate answer and rationale}
            \IF{$A' = A$}
                \STATE Record $\{Q, C, A', r\}$ with critique label \texttt{Accept} \label{alg:line-accept}
            \ELSE
                \STATE Record $\{Q, C, A', r\}$ with critique label \texttt{Reject} \label{alg:line-reject}
            \ENDIF
            \STATE ${t} \gets {t} + 1$

            \STATE $q_t \gets \mathcal{R}(Q, C)$ \label{alg:line-generatequery} \hfill \textcolor{gray}{// generate retrieval query}
            \STATE $d_t \gets \mathcal{IR}(q_t)$ \label{alg:line-retrieval} \hfill \textcolor{gray}{// retrieve document}
            \STATE $C \gets C \cup \{q_t, d_t\}$ \label{alg:line-updatecontext} \hfill \textcolor{gray}{// update context}

        \ENDWHILE
    \ENDFOR
    
    \end{algorithmic}
    \end{algorithm}

Our training pipeline starts with collecting training data for the supervised learning of the Critic. 
Given the Critic's task, the training data should consist of information-seeking chains with accurate labels and include sufficient quantity and diversity, enabling the Critic to learn how to assess mid-step information sufficiency and evaluate the correctness of the current answer in real-world scenarios.
However, as discussed in Section~\ref{sec:intro}, information sufficiency is subject to the LLM's knowledge and other available information. Thus, human-labeled, model-agnostic information-seeking chains may not match the real information-seeking behavior of an LLM whose internal knowledge scope does not align with the annotator's. 
To address this, we propose an approach to generate model-specific and context-aware synthetic data instead of real human-annotated data. 
Specifically, we let the RAG system self-practice a multi-round retrieval process to find the correct answer for a given question where the target answer is known. During this practice, the real interaction between the Reasoner and the Retriever enables us to collect and label inner monologue data. 

As shown in Algorithm~\ref{alg:data_generation}, the Self-Practicing Algorithm requires only a source dataset containing the initial question $Q$ and the final answer $A$, which are readily available in most existing QA datasets. The augmented information is tracked by the context list $C$, initialized as an empty list (line~\ref{alg:line-initialization}). For each data point, an iterative process simulates inner monologue between the Reasoner $\mathcal{R}$ and search engine $\mathcal{IR}$. In the $t$-th round, given all currently available information (i.e., embedded knowledge of LLM and the augmented retrieved information $C$), the Reasoner generates an answer $A'$ and rationale $r$ for the question $Q$ (line~\ref{alg:line-generateanswer}). A correctness check between $A$ and $A'$ determines whether the critique label $y$ for the data tuple $x=\{Q,C,A',r\}$ should be \texttt{Accept} or \texttt{Reject} (line~\ref{alg:line-accept} and \ref{alg:line-reject}). Regardless of the label, one data point is generated, and the process continues. For each raw ($Q$, $A$) pair, the algorithm generates $T$ training data points of varying chain lengths and labels, where $T$ is a predefined maximum number of turns (i.e., rounds of answering).
If the exit condition is unmet, the Reasoner is re-prompted with $Q$ and $C$ to generate a search query $q_t$ (line \ref{alg:line-generatequery}). The search engine  $\mathcal{IR}$ is then called to retrieve passages/documents $d_t$ given the query $q_i$ (line \ref{alg:line-retrieval}). Next, the context is updated by concatenating the new query and document, $C = C \cup \{q_t, d_t\}$, before the next iteration (line \ref{alg:line-updatecontext}).

As a result, given a source dataset with $N$ question-answer pairs, the Self-Practicing Algorithm generates an augmented training set $\mathcal{D} = \{(x_i, y_i)\}_{i=1}^{M}$, where $M={N\times T}$ is the size of $\mathcal{D}$. 
The generated data is annotated with current answer correctness (\texttt{Accept} or \texttt{Reject}) and entails information sufficiency for a given LLM under a multi-round RAG setting. Note that each generated data point simulates a sequence of a RAG system's behavior, with either positive or negative critique labels. This differs from standard LLM knowledge distillation methods, which treat all distilled data points as positive training examples. Moreover, the same raw ($Q,A$) pair can generate different sequences of RAG behaviors and corresponding labels across iterations when the available information changes. This increases both the quantity and diversity of the synthetic data.

\subsection{Reasoning-Enhanced Critic Learning}
\label{sec:training}
The Critic’s task is formulated as a straightforward binary classification problem for a given input $x$. While any classification algorithms could be used, we use pre-trained language models in this work due to their superior semantic understanding capabilities. Thus, the Critic training is to tune a generative language model by casting the classification task as a conditional generation problem.

Given the training set $\mathcal{D} = \{(x_i, y_i)\}_{i=1}^{M}$ from Section~\ref{sec:data-gen},
we follow the conception of instruction fine-tuning and first update the input $x_i$ to be the concatenation of general language instruction and task-specific input. During the training, the Critic is expected to predict the next token $y_i$ given input $x_i$; thus, the training objective is to maximize the conditional likelihood of the binary labels (i.e. targeted tokens), as shown in Equation~\ref{eq:loss}, where $\theta$ is the language model parameters.

\begin{equation}
\theta^* = \arg\max_\theta \log \mathcal{L}(\theta)= \arg\max_\theta \sum_{(x_i, y_i) \sim \mathcal{D}} \left[ \log P(y_i | x_i; \theta) \right]
\label{eq:loss}
\end{equation}

Through task-specific fine-tuning, the Critic quickly learns to make predictions from a large amount of labeled IM-distilled data.

\subsection{Reasoning-Enhanced Inference}
\label{sec:inference}
The goal of this section is to justify and elaborate on how the mechanism introduced in Sections \ref{sec:data-gen} and \ref{sec:training} improves RAG's output by leveraging the Critic’s meta-cognitive feedback.
The integration of the trained Critic with the Reasoner and Retriever transforms SIM-RAG into an iterative reasoning framework, enabling it to dynamically refine responses based on feedback. The supervision of the LLM-based Reasoner over multiple rounds is verbal Reinforcement Learning (RL), where the Critic provides supervision in natural language (\texttt{Accept} or \texttt{Reject}) rather than numerical rewards or gradient updates. Analogous to traditional policy-based RL setups, this verbal reinforcement defines a policy as a combination of the agent’s memory encoding and the selected LLM parameters~\cite{shinn2024reflexion}. 
Within our framework, this approach leverages the strengths of in-context learning~\cite{brown2020language}, as the Reasoner can interpret and incorporate the Critic’s feedback directly into its reasoning process by appending it to the input of the Reasoner. 
This in-context RL mechanism enables the Reasoner to dynamically adapt its behavior and decision-making process based on $Q$ and $C$, including the supervision provided by the Critic, without requiring explicit parameter updates. This preserves the system's modularity and keeps training lightweight. By grounding iterative refinement in textual feedback, the framework encourages targeted improvements in reasoning paths and retrieval strategies.

From the system design perspective, SIM-RAG separates the Critic from the Reasoner to avoid self-critique bias. During inference time, the iterative cycle of feedback and context updates mirrors human reasoning under uncertainty, where gaps in knowledge are identified and addressed step-by-step.

\section{Experiments}

\subsection{Task and Datasets}
To comprehensively evaluate SIM-RAG across tasks of varying reasoning complexity, we conduct experiments on three highly distinct datasets covering single-hop and multi-hop QA tasks.
For single-hop QA, we use TriviaQA~\cite{joshi2017triviaqa}, a widely used benchmark that focuses on factoid questions requiring reasoning over a single piece of evidence from Wikipedia. 
For multi-hop QA, we use HotpotQA~\cite{yang2018hotpotqa} and 2WikiMultiHopQA~\cite{ho2020constructing}. HotpotQA requires synthesizing information from multiple documents to answer complex questions, while 2WikiMultiHopQA focuses on distinguishing closely related entities and incorporating fine-grained evidence. These datasets challenge SIM-RAG’s capabilities in multi-round retrieval and reasoning over insufficient information. Following the standard evaluation methods, we report Exact Match (EM) and F1 scores for all datasets, using the Wikimedia dumps provided by each dataset as the retrieval corpus.

\subsection{Implementation Details}
We evaluate two versions of SIM-RAG using Llama3-8B and GPT-4 as the Reasoner. To fine-tune the Critic, we use two Flan-T5 models of different sizes, corresponding to two versions of SIM-RAG: Flan-T5-2.85B for the full version (SIM-RAG\textsubscript{\textit{full}}) and Flan-T5-783M for the lightweight version (SIM-RAG\textsubscript{\textit{lite}}).
To ensure consistency with other well-known RAG frameworks, we use BM25 with Elasticsearch as the retriever across all experiments. Our pipeline can be replicated with two NVIDIA 3090 GPUs or equivalent hardware. We consider prompts, in-context examples, and the number of documents retrieved as hyperparameters and report them in our open-sourced code base to facilitate reproducibility.

\subsection{Baselines}
\label{sec:baselines}

\begin{table}[tb]
    \centering
    \resizebox{\columnwidth}{!}{
    \begin{tabular}{p{0.15cm} p{2.1cm} p{1.9cm} cccccc}
        \toprule
         & & & \multicolumn{2}{c}{TriviaQA} & \multicolumn{2}{c}{HotPotQA} & \multicolumn{2}{c}{2Wiki}\\
        \cmidrule(r){4-5}\cmidrule(r){6-7}\cmidrule(r){8-9}
         & Method & LLM/Retriever & EM & F1 & EM & F1 & EM & F1\\
        \midrule
        \multirow{8}{*}{{\color{gray}{(a)}}}
        & \multicolumn{8}{c}{\footnotesize{\textit{Prompting}}}\\
        & NaiveGen~\cite{jin2024flashrag}                                              & Llama3            & 55.7 & 63.1 & 20.0 & 28.4 & 26.4 & 33.9\\
        & StandardRAG~\cite{jin2024flashrag}                                           & Llama3/E5         & 58.9 & 68.3 & 25.1 & 35.3 &  8.6 & 21.0\\
        & SEAKR~\cite{yao2024seakr}                              & Llama2/BM25       & 42.6 & 48.6 & 27.9 & 39.7 & 30.2 & 36.0\\
        & DRAGIN~\cite{su2024dragin}                             & Llama2/BM25       & 54.4 & 62.3 & 23.7 & 34.2 & 22.4 & 30.0\\
        \cmidrule(lr){2-9}

        & \multicolumn{8}{c}{\footnotesize{\textit{Learned}}}\\
        & Self-RAG~\cite{asai2023self}*                           & Llama3/E5         & 38.2 & 53.4 & 17.1 & 29.6 & 12.1 & 25.1\\
        & Auto-RAG~\cite{yu2024auto}*                             & Llama3/E5         & 56.7 & 60.7 & 31.9 & 44.3 & 38.8 & 46.9\\
        \midrule\midrule
        \multirow{10}{*}{{\color{gray}{(b)}}}
        & \multicolumn{8}{c}{\footnotesize{\textit{Prompting}}}\\
        & NaiveGen~\cite{jin2024flashrag}                                                & GPT-4             & 59.6 & 70.0  & 27.5   & 37.9 & 23.7  & 31.7\\
        & StandardRAG~\cite{jin2024flashrag}                                             & GPT-4/BM25          & 60.3 & 65.9  & 35.7 & 45.5 & 28.7   & 33.8\\
        & FLARE~\cite{jiang2023active}*                           & Llama3/E5         & 55.8 & 63.2 & 19.7 & 28.0 & 25.8 & 33.9\\
        & IR-CoT~\cite{trivedi2023interleaving}*                   & Llama3/BM25         & 56.4 & 68.9 & 28.6 & 39.1 & 23.5 & 31.8\\
        \cmidrule(lr){2-9}
        & \multicolumn{8}{c}{\footnotesize{\textit{Learned}}}\\
        & SIM-RAG\textsubscript{\textit{lite}}                                        & Llama3/BM25       & 69.2 & 74.5 & 27.8 & 37.9 & 32.1 & 38.0\\
        & SIM-RAG\textsubscript{\textit{lite}}                                       & GPT-4/BM25        & 77.3 & 82.8 & 37.0 & 49.7 & 44.5 & 53.3\\
        & SIM-RAG\textsubscript{\textit{full}}                                             & Llama-3/BM25       & 70.7 & 75.6 & 32.7 & 43.3 & 34.1 & 40.2\\
        & SIM-RAG\textsubscript{\textit{full}}                                            & GPT-4/BM25        & 77.5 & 82.7 & 39.8 & 52.2 & 46.1 & 54.6\\
        \bottomrule
    \end{tabular}}
    \caption[]{Comparison of RAG-based methods on three standard RAG datasets. Group (a): methods that require direct access to model weights; Group (b): methods that can be used with API-based LLMs\setcounter{footnote}{2}\footnotemark. Asterisk (*) indicates the results obtained from recent reproduction using newer models.}
    \label{tab:results}
\end{table}

We compared SIM-RAG with eight baseline methods (Table \ref{tab:results}). Naive Generation ~\cite{jin2024flashrag} and Standard RAG ~\cite{jin2024flashrag} are two basic methods. Naive Generation fully relies on the internal knowledge of the LLM without using any retrieval, whereas Standard RAG uses the initial question as a query to retrieve documents, which are then used to augment the LLM’s response through prompting. We report the Llama3 and GPT-4 version baselines for both methods as a reference point for performance comparison. SEAKR~\cite{yao2024seakr}, DRAGIN~\cite{su2024dragin}, Self-RAG~\cite{asai2023self}, Auto-RAG~\cite{yu2024auto}, FLARE~\cite{jiang2023active}, IR-COT ~\cite{trivedi2023interleaving} are more advanced multi-round RAG methods. Some baselines are prompt-based, and others are learned.

\footnotetext{While Naive Generation and Standard RAG would technically fit in Group (b) (because they can be used via API), we list their Llama-based variants in Group (a) for easier comparison with other Llama-based methods.}

The baselines include two groups: (a) methods requiring access to an LLM's internal weights and open-source models, and (b) methods that can use API-based closed-source models.
In Group (a), SKEKR~\cite{yao2024seakr} and DRAGIN~\cite{su2024dragin} do not involve fine-tuning; however, they rely on model internals, such as hidden-layer Gram matrices (SKEKR) or token-level attention weights (DRAGIN), for retrieval. Self-RAG~\cite{asai2023self} and Auto-RAG~\cite{yu2024auto} fine-tune LLMs to support multi-round retrieval. 
While Self-RAG originally used Llama2, we report results based on its recent replication~\cite{jin2024flashrag} using Llama3 for a fair comparison. 
Compared to SKEKR, DRAGIN, Self-RAG, and Auto-RAG methods, which require access to model weights, SIM-RAG offers greater flexibility without requiring open-source models.

Group (b) focuses on methods applicable to API-based closed-source LLMs, primarily relying on prompting. FLARE~\cite{jiang2023active} utilizes probabilities or confidence scores of the next-generated-tokens to guide retrieval, while IR-CoT~\cite{trivedi2023interleaving} interleaves retrieval with intermediate CoT reasoning steps, enabling effective multi-step question answering. For better comparison, we include replication results from~\cite{jin2024flashrag} using Llama3. 
Baselines with E5-base-v2~\cite{wang2022text} as the Retriever potentially have an advantage due to higher-quality retrieval engines; however, they also require more GPU resources.

\subsection{Results}
\label{sec:results}

Table~\ref{tab:results} summarizes the performance of various methods on three widely used RAG datasets. 
SIM-RAG\textsubscript{\textit{full}} with GPT-4 consistently performs the best in all three datasets, outperforming all baseline methods by a large margin, including Self-RAG and Auto-RAG, which require extensive full model fine-tuning for LLMs. A group-wise comparison further highlights the strengths of SIM-RAG. 

For multi-hop QA datasets, when using closed-source GPT-4 models, SIM-RAG\textsubscript{\textit{full}} delivers the highest performance for both HotPotQA and 2Wiki. When using open-source Llama models, Auto-RAG performs the highest on 2Wiki, while SIM-RAG\textsubscript{\textit{full}} is the highest on HotPotQA. Auto-RAG fine-tunes Llama3-8b for retrieval decisions and uses a learned retrieval E5, whereas our approach only fine-tunes a smaller Critic and uses a simpler BM25 retriever. If reducing computational costs or using closed-source LLM models is a priority, SIM-RAG would be the optimal choice. 

In contrast, for single-hop QA, a key observation is that all advanced baselines tailored for multi-round RAG perform poorly on the straightforward TriviaQA dataset. None of them match the performance of Standard RAG, and only Auto-RAG and FLARE outperform Naive Generation.
This exposes a critical limitation for these approaches: optimizing for complex multi-round retrieval tasks appears to undermine LLM's capability on simpler tasks. 
This may be due to the inherent biases of LLMs, particularly their difficulty with effective self-critique, as discussed in Section~\ref{sec:mutliRoundRAG}.  
These challenges lead to Over-Confidence or Over-Retrieval, making LLMs less competitive on simpler tasks, where even a standard RAG approach with a fixed number of retrieval steps performs better.
In contrast, SIM-RAG uses an external model that specializes in ``when to stop the iterative system.'' This distinction allows SIM-RAG\textsubscript{\textit{full}} with Llama3 to outperform Standard RAG with Llama3 by a significant margin (70.7\% vs. 58.9\%) in the EM metric. Similarly, SIM-RAG\textsubscript{\textit{full}} with GPT-4 achieves a 16.0\% absolute improvement over the Standard RAG baseline, representing a relative improvement of 26.1\%.

The study on Critic size further reveals that SIM-RAG is an effective framework even with a lightweight Critic (783M). For example, on HotPotQA, SIM-RAG\textsubscript{\textit{lite}} significantly outperforms Self-RAG (27.8\% vs. 17.1\%) while using only one-tenth of the training parameters (783M vs. 7B).

These findings indicate that even a lightweight Critic can improve system performance, though SIM-RAG may benefit from a larger and more powerful Critic, particularly for complex multi-hop reasoning tasks.

\subsection{Analysis on Critic Predictions}

\begin{figure}
        \centering
        \includegraphics[width=\linewidth]{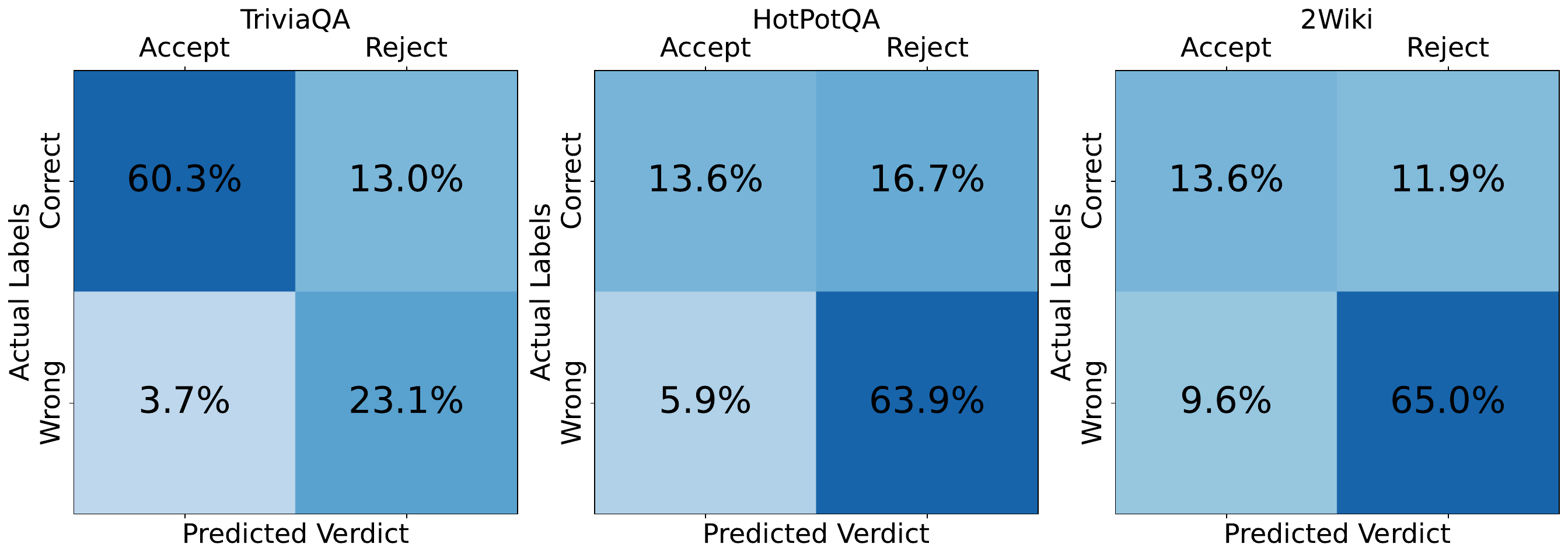}
        \caption{Confusion matrices of Critic predictions in SIM-RAG with GPT-4.}
        \Description[]{}
        \label{fig:confusion-matrix}
    \end{figure}

To further evaluate the Critic’s prediction accuracy, we report SIM-RAG with GPT-4's confusion matrices of its binary classification performance on TriviaQA, HotPotQA, and 2Wiki datasets (Figure~\ref{fig:confusion-matrix}). The clear diagonal lines (True Positive and True Negative) highlight the Critic’s ability to correctly predict whether to accept or reject a Reasoner’s output, based on the ground-truth answer. From the results, the Critic demonstrates strong classification performance across all datasets, with notably high accuracy in rejecting incorrect answers, particularly for HotPotQA (63.9\%) and 2Wiki (65.0\%). 

However, the True Positive and True Negative rates differ significantly between single-hop and multi-hop QA tasks. On TriviaQA, the Critic achieves 60.3\% accuracy in correctly accepting answers, while on HotPotQA and 2Wiki, the accuracy drops to 13.6\%.  These discrepancies are expected and reflect the differing characteristics of the datasets. SIM-RAG demonstrates an ability to adapt across datasets, likely due to the distribution of the synthetically generated training data, which contains a higher proportion of positive examples in TriviaQA compared to the multi-hop datasets.

\subsection{Ablation Study and Analysis}
\label{sec:ablation}

\paragraph{Impact of Critic Model Choice}

\begin{figure}
        \centering
        \includegraphics[width=\linewidth]{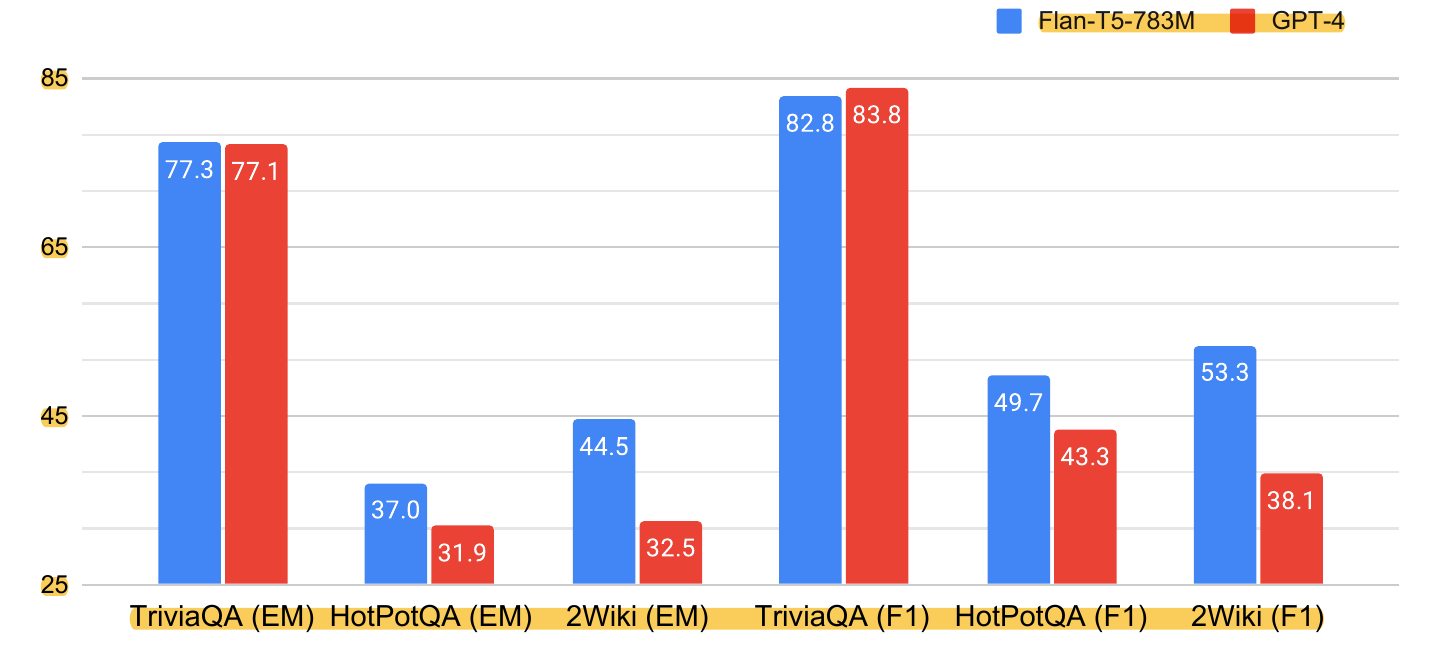}
        \caption{Ablation study on different Critic model choices. Both follow the SIM-RAG setting with GPT-4 as the Reasoner, but the blue bar uses learned Flan-T5-783M as a Critic, while the red bar uses GPT-4 as the Critic (a.k.a., self-critique).}
        \Description[]{}
        \label{fig:self-critic-ablation}
    \end{figure}

We conducted an ablation study to explore whether a more powerful, general-purpose model with strong QA capabilities could serve as a better Critic. To this end, we replaced Flan-T5-783M with GPT-4 as the Critic in our SIM-RAG\textsubscript{\textit{lite}} system, while using GPT-4 as the Reasoner. In this configuration, GPT-4 functions as both the Reasoner and Critic using different prompt settings, an approach commonly referred to as self-critique in literature~\cite{gou2023critic}. The comparison is visualized in Figure~\ref{fig:self-critic-ablation}. Notably, unlike other baselines discussed in Table~\ref{tab:results}, GPT-4-as-the-Critic achieves strong results on TriviaQA. This suggests that for simple tasks, SIM-RAG may use a more general, powerful model like GPT-4 to achieve comparable outcomes. 
However, the gap becomes significant for more complex multi-hop tasks. As shown in the bar chart, GPT-4-as-the-Critic underperforms Flan-T5 across both EM and F1 metrics by a substantial margin. This finding aligns with observations in the mathematical reasoning domain~\cite{huang2024large}, where LLMs often struggle to provide reliable self-critiques on tasks involving complex reasoning. 
We suspect that a general-purpose LLM may be over-confident and generate too many false positives as the Critic. This is a more serious problem for multi-round QA tasks, where the percentage of actual ``correct'' predictions from the Reasoner is low (as suggested by the confusion matrices in Figure \ref{fig:confusion-matrix}).

\paragraph{Analysis of Inner Monologue Turns}

    \begin{figure}[h]
        \centering
        \includegraphics[width=0.95\linewidth]{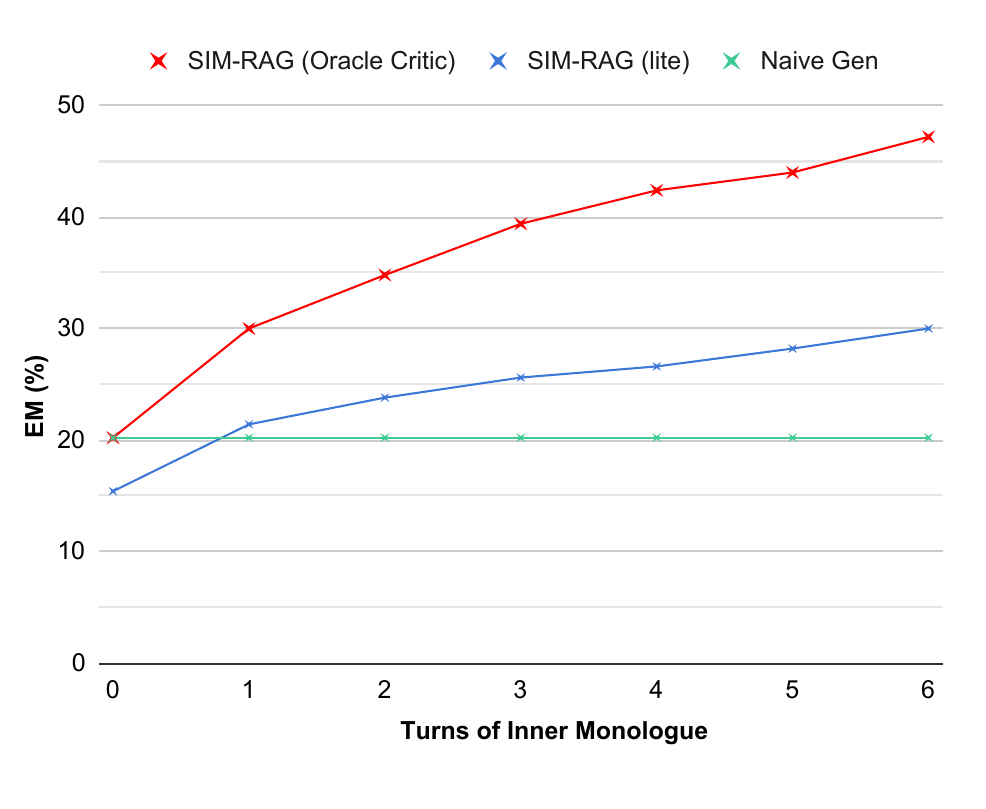}
        \caption{Ablation study on different Inner Monologue turns on HotPotQA. The green line represents the Naive Generation baseline, while the blue line shows the performance of SIM-RAG\textsubscript{\textit{lite}}. The red line indicates the EM score using the ground-truth 
        label as an Oracle Critic.}
        \Description[]{}
        \label{fig:max-turns}
    \end{figure}

As described in Section~\ref{sec:sim-rag-main}, the SIM-RAG Self-Practicing Algorithm (Algorithm \ref{alg:data_generation}) defines an arbitrary maximum number of retrieval turns $T$ during self-practice and inference. In practice, however, the number of turns is typically kept relatively low to strike a balance between the cost of inference (in terms of time and computational resources) and performance, while also taking into account the inherent limitations of LLM capabilities (more details in Section~\ref{sec:discussion}).
To better understand the impact of the hyperparameter $T$, we conducted an ablation study varying $T$ from 0 to 6. To save computational time, this experiment was performed on a subset of the HotPotQA development set, comprising 500 questions from the previously established RAG evaluation benchmark~\cite{zhang2024raglab}.

Figure~\ref{fig:max-turns} shows that larger $T$ leads to better performance, and there is much room to improve the performance of SIM-RAG by optimizing this hyperparameter. The green line represents the Naive Generation baseline, providing a reference point on LLM's zero-shot performance. The blue line shows SIM-RAG\textsubscript{\textit{lite}}’s actual performance across different turn settings, reflecting the system’s ability to refine answers iteratively. Finally, the red line indicates the performance ceiling achievable using an Oracle Critic, which has access to the ground-truth label to determine whether to accept the current answer as the system’s output.

By the final round, SIM-RAG achieves a 10.1\% improvement over Naive Generation, corresponding to a 50.5\% relative gain. Notably, using the Oracle Critic achieves 47.2\%, representing the theoretical upper limit of SIM-RAG without modifying the Reasoner. This highlights the potential for further improvements in SIM-RAG through a better or larger Critic, as discussed in Section~\ref{sec:results}.

When $T=0$, SIM-RAG yields fewer exact-match answers than Naive Generation, primarily because the Critic occasionally rejects correct answers. This outcome is expected, as SIM-RAG is designed to abstain from providing answers when evidence is insufficient, thereby reducing hallucinations (i.e., false positives) at the expense of occasionally suppressing correct responses (i.e., true positives or exact-match answers). We limit $T$ to 6 in our experiments due to the context length constraint of the Critic. However, with ongoing advances in extending language model context windows, this limitation is expected to become less restrictive in the future.

\paragraph{Analysis on Self-Practicing Data}
Figure~\ref{fig:teaser} visualizes the inner monologue traces generated during self-practicing. These traces contribute to a rich and diverse training set, which naturally covers both insufficient and excessive retrieval cases. In particular, the self-practicing data captures how retrieval decisions influence reasoning outcomes. As illustrated in the four examples, the 0-Round and 1-Round traces show incorrect predictions due to missing evidence, while the 2-Round trace demonstrates a correct prediction when sufficient information is retrieved. Moreover, the 3-Round trace shows how over-retrieval introduces irrelevant information, distracting the model from correct prediction.
Unlike humans, who can selectively attend to relevant information, LLMs are highly sensitive to noisy~\cite{shi2023large}, misleading~\cite{zeng2025worse}, or even long~\cite{liu2024lost} inputs. Consequently, retrieving more information is not always beneficial for LLMs. This characteristic highlights the importance of information sufficiency checking and explains why self-practicing data is valuable in SIM-RAG.

\section{Discussion}
\label{sec:discussion}
\subsection{Strengths and Weaknesses}

\paragraph{Task}

As shown in Section~\ref{sec:results}, SIM-RAG excels in handling RAG tasks of varying complexity, from simple single-round retrieval to multi-round reasoning tasks. Its flexibility enables consistent outperformance of traditional baselines.

\paragraph{Domain adaptation}
Like most training-free domain adaptation approaches for LLMs, SIM-RAG performs well on tasks that align with the Reasoner’s pretraining corpus but may face challenges with domain-specific jargon or highly specialized terminology. The Critic, on the other hand, is system-specific. The behavior-driven training (discussed in Section~\ref{sec:Critic-learning}) ensures it is well aligned with the system used to generate the synthetic training data (in Algorithm \ref{alg:data_generation}). However, if any major component of a RAG system, such as the Reasoner or Retriever. is replaced or significantly updated, the Critic may require retraining to maintain optimal performance. 

\paragraph{Computational cost}
The computational cost of SIM-RAG comprises two main components: training and inference. For training, the primary computational expense comes from data generation and Critic learning. The data generation phase requires $(T \times N)$ large model inferences, where $T$ represents the predefined maximum number of retrieval rounds, and $N$ denotes the size of the source dataset. The Critic training phase follows the standard resource demands of supervised learning. During inference, efficiency depends on the Reasoner’s capabilities. If the question aligns with the LLM's pretrained knowledge, SIM-RAG is efficient. However, for unfamiliar domains, SIM-RAG may require more turns, highlighting the importance of domain adaptation to optimize performance and reduce inference time.

\paragraph{Failure Cases}

Figure \ref{fig:teaser} provides an example of how our system can both benefit from and be hindered by the Critic rejecting responses, as false rejections may lead to mistakes due to Over-Retrieval in the 3-Round example. Beyond the Critic’s behavior, as shown in Figure \ref{fig:max-turns}, even an Oracle Critic yields low accuracy, highlighting that failures also stem from broader limitations. These include the dataset’s inherent difficulty, which challenges both the Reasoner’s knowledge and its ability to formulate effective search queries, as well as the quality of the Retriever.

\subsection{Critic Learning}
\label{sec:Critic-learning}
From our empirical study, we found that introducing the Critic is a simpler problem than solving the task. 
Prior studies show that Flan-T5-783M (the Critic in SIM-RAG\textsubscript{\textit{lite}}) struggles to handle complex tasks through direct fine-tuning on training datasets~\cite{li-etal-2024-teaching}. Flan-T5-783M achieves only 14.7\% EM score on the HotPotQA dataset via fine-tuning, compared to 20.1\% achieved by zero-shot Llama3. However, Flan-T5-783M can be trained in SIM-RAG to do a decent job of serving as a Critic for more powerful LLMs (Llama3 and GPT-4). One possible reason is that the Critic only needs to model the relationships between the question, query, retrieved documents, the LLM’s predicted answer and rationale, and the correctness of that prediction. That is, the Critic does not necessarily ``know'' the correct answer or how to generate the correct answer; instead, it has the easier task of learning to evaluate the coherence and validity of the LLM’s output.

\section{Conclusion and Future Work}

In this paper, we propose the SIM-RAG framework, a lightweight yet effective approach to optimizing multi-round RAG by adding a critic module. This framework can work with both close-source and open-source RAG systems, offering great flexibility in real-world applications. 
Since SIM-RAG does not modify the parameters of the LLM, it functions as an inference-time enhancement to the RAG system, similar in goal to OpenAI-o1, but realized through a fundamentally different mechanism.
In particular, this work introduces a novel Self-Practicing Algorithm to generate synthetic data to address the shortage of labeled, system-specific multi-round reasoning and retrieval training data. Experiments on three standard benchmarks validate SIM-RAG’s effectiveness.

This work opens up additional opportunities for future enhancement of generative AI. Although SIM-RAG is a lightweight approach without requiring access to an LLM’s weights, we recognize the potential for the trained Critic to be used as a reward model in future work to optimize other components of RAG (Retriever, Reasoner, etc.) through policy-based actor-critic reinforcement learning (e.g., RLHF). Moreover, the capability of an AI system to recognize its limitations—\textit{Knowing you don't know}—is crucial for reducing hallucination and enhancing trustworthiness and reliability. While this paper focuses on multi-round RAG, we expect the novel Self-Practicing and Critic training techniques to be widely adopted to other AI problems.

\section*{Acknowledgment}
This work was supported by the IRKM Lab at the University of California, Santa Cruz. Special thanks to Ke Yang for the valuable discussions, to Jeshwanth Bheemanpally and Li Liu for their constructive feedback, and to Linda Li for the visualizations of the system and data.

\bibliographystyle{ACM-Reference-Format}
\bibliography{arxiv-submission}


\end{document}